\documentclass[sigconf]{acmart}

\usepackage{balance}

\def\BibTeX{{\rm B\kern-.05em{\sc i\kern-.025em b}\kern-.08emT\kern-.1667em\lower.7ex\hbox{E}\kern-.125emX}}

\usepackage{listings}
\usepackage{amsmath}
\copyrightyear{2020}
\acmYear{2020}
\setcopyright{usgovmixed}
\acmConference[ICMR '20]{Proceedings of the 2020 International Conference on Multimedia Retrieval}{June 8--11, 2020}{Dublin, Ireland}
\acmBooktitle{Proceedings of the 2020 International Conference on Multimedia Retrieval (ICMR '20), June 8--11, 2020, Dublin, Ireland}
\acmPrice{15.00}

\begin{document}

\fancyhead{}

\title{HLVU : A New Challenge to Test Deep Understanding of Movies the Way Humans do}

\author{Keith Curtis}
\affiliation{%
  \institution{National Institute of Standards and Technology}
  \city{Gaithersburg}
  \state{Maryland}
  \country{USA}
}
\email{keith.curtis@nist.gov}

\author{George Awad}
\affiliation{%
  \institution{National Institute of Standards and Technology}
  \city{Gaithersburg}
  \state{Maryland}
  \country{USA}
}
\authornotemark[1]
\email{george.awad@nist.gov}

\author{Shahzad Rajput}
\affiliation{%
  \institution{National Institute of Standards and Technology}
  \city{Gaithersburg}
  \state{Maryland}
  \country{USA}
}
\authornote{Georgetown University}
\email{shahzad.rajput@nist.gov}

\author{Ian Soboroff}
\affiliation{%
  \institution{National Institute of Standards and Technology}
  \city{Gaithersburg}
  \state{Maryland}
  \country{USA}
}
\email{ian.soboroff@nist.gov}

%
\renewcommand{\shortauthors}{Curtis et al.}

%
\begin{abstract}
  In this paper we propose a new evaluation challenge and direction in the area of High-level Video Understanding. The challenge we are proposing is designed to test automatic video analysis and understanding, and how accurately systems can comprehend a movie in terms of actors, entities, events and their relationship to each other. A pilot High-Level Video Understanding (HLVU) dataset of open source movies were collected for human assessors to build a knowledge graph representing each of them. A set of queries will be derived from the knowledge graph to test systems on retrieving relationships among actors, as well as reasoning and retrieving non-visual concepts. The objective is to benchmark if a computer system can "understand" non-explicit but obvious relationships the same way humans do when they watch the same movies. This is long-standing problem that is being addressed in the text domain and this project moves similar research to the video domain. Work of this nature is foundational to future video analytics and video understanding technologies. This work can be of interest to streaming services and broadcasters hoping to provide more intuitive ways for their customers to interact with and consume video content.
\end{abstract}

%
%


%

%
\maketitle

\section{Introduction}
Video understanding is a very difficult problem to solve. All current video analysis technology relies on detection, recognition and analysis of certain specific visual concepts such as people, objects, actions, activities or events. All these visual concepts recognition are usually done in isolation from the context of the video and only focusing on visual clues. This is one of the reasons why the current state of the art is lacking high level video understanding capabilities that connect the entities of a video with different events or relations. For example, given a two hour movie the current computer vision systems are not able to understand the relationship between different characters and develop a deep understanding of the video context. There has been efforts to encourage research in high level video understanding such as the “MovieQA" and "The Large Scale Movie Description Challenge" \cite{LSMDC}. However these tasks revolve around isolated visual concepts retrieval and not about testing systems for their overall understanding of entities, relations and events within the video/movie. Early visions of this kind of work \cite{debattista2018expressing} proposed to use visual and audio descriptors, in addition to employing semantic analysis and linking with external knowledge sources in order to populate a knowledge graph. 

The goal of our proposed research is to design and build datasets and evaluation benchmarks to foster the interest of the research community to develop systems which can extract available information from a video (e.g. a movie characters, their story lines, and relationships), and to use this information to reason about other, more hidden background information, and eventually to populate a knowledge graph with all extracted and reasoned information. In the next section we present an overview to related work in this area followed by more technical details discussing a pilot dataset collection of open source movies, the human annotation framework, query design and preprocessing, and our proposed evaluation roadmap and future directions. This is an ambitious new area of research which is to be run initially as a Grand Challenge at ACM Multimedia 2020 and as a workshop at ACM ICMI 2020 respectively. TRECVID workshop participants will be invited to participate in this new task.

\section{Related Work}
Integrating vision and language research has recently gained a lot of attention to promote image and video understanding. Most approaches adopt the question answering paradigm as their evaluation framework. For example in \cite{antol2015vqa} the authors propose the task of visual question answering for images. They provide a dataset containing ~0.25M images, ~0.76M question and ~10M answers. However, most question types target very specific visual facets in the image such as what, where, number of objects, and what is the attributes or relation of an object to others.

MovieQA \cite{tapaswi2016movieqa} is a dataset which aims to evaluate automatic story comprehension from video and text. It consists of 14,944 multiple choice questions, each with 5 multiple-choice answers, only one of which is correct, from about 408 movies with high semantic diversity. The movies have been segmented into video clips of maximum 200s durations where participants have to answer a question related to this video clip. The dataset itself comes with multiple answering sources for the questions such as plot synopses, scripts, subtitles, and audio descriptions. Annotators essentially used the plot synopses to come up with the set of questions and answers instead of watching the whole movie.

Following from this, \cite{jasani2019we} explored the biases in the MovieQA dataset and found that by using an appropriately trained word embedding, about half of the Question-Answers can be answered by looking at the questions and answers alone, completely ignoring the narrative context from video clips, subtitles, and movie scripts.

A large-scale dataset of corresponding movie trailers, plots, posters, and metadata was developed by \cite{cascante2019moviescope} who study the effectiveness of visual, audio, text, and metadata-based features for predicting high-level information about movies such as their genre or estimated budget.

The large-scale movie description challenge \cite{LSMDC} was first held as a workshop at the International Conference on Computer Vision (ICCV) 2015. This was held as a unified challenge on Text generation using single video clip, and Text generation using single video clip as well as its surrounding context. This challenge was later also held at ICCV 2017. It was held as two challenges: The large scale movie description and understanding challenge [LSMDC] and [MovieQA]. The LSMDC task included Movie description, Movie annotation and retrieval, and Movie fill-in-the-blank task. The MovieQA task included Question-answering in movies and video retrieval based on plot synopses sentences. A follow-up challenge was also held at ICCV 2019.

Early visions of the proposed work \cite{debattista2018expressing} explored the usage of visual and audio descriptors, in addition to employing semantic analysis and linking with external knowledge sources in order to populate a knowledge graph. Under this approach, time-stamped multi-modal signals of the video would be uplifted into a format that could be utilised by the applications semantic model and the resulting knowledge graph could be searchable in a newly meaningful way.

ActivityNet \cite{caba2015activitynet} is a large scale video benchmark for human activity understanding. ActivityNet provides samples from 203 activity classes with an average of 137 untrimmed videos per class and 1.41 activity instances per video, for a total of 849 video hours. All ActivityNet videos are obtained from online video sharing websites. Amazon Mechanical Turk (AMT) workers are used to determine if videos contain the intended activity class. AMT workers also label the beginning and end points of each activity. 

The How2 dataset \cite{sanabria2018how2} was introduced by Sanabria et al. This is a large scale multimodal language understanding dataset to aid in the development  human like language understanding capabilities, where machines should be able to jointly process multimodal data, and not just text, images, or speech in isolation.

The TRECVID Instance Search (INS) \cite{2019trecvidawad} task is a benchmarking task run at NIST (U.S. National Institute of Standards and Technology) to measure systems performance on the retrieval of specific instances of either persons, locations, or objects. In the last three years the task targeted pairs of instances such as the retrieval of video shots of specific people in given locations within the British Broadcasting Corporation (BBC) Eastenders TV series. While recently the task targeted video shots of specific people doing specific actions in the Eastenders world. 

TRECVID Video to Text (VTT) \cite{2019trecvidawad} is a benchmarking task also run at NIST to evaluate the performance of systems that automatically generate a single sentence description for short videos. Videos are typically shorter than 10 seconds in length and are taken from social media to represent real world situations. Although ActivityNet, INS and VTT benchmarks are all important efforts to encourage research in recognition and video semantics, they are still considered component tasks and not targeting the real holistic problem of high-level video understanding the way humans do.

\section{Technical Details}
\subsection{Dataset}
The procurement of suitable datasets is vital for the undertaking of this new research area. The authors have spent time identifying Movies with a Creative Commons (CC) license \cite{CC} which can be disseminated to participating researchers for the purpose of this research. The most important criteria in selecting the movies were reasonable video quality, duration of more than 15 min at least, and self contained story lines with clear actors, relations, events and entities. In total, a pilot dataset of about 11\thinspace hrs has been collected from public websites such as Vimeo\footnote{https://vimeo.com/} and the Internet Archive\footnote{https://archive.org/}.

Table \ref{dataset} shows the current set of collected movies, their genre and durations. All movies have been deemed by the authors to be suitable for this research, and will be disseminated to participants as appropriate. The main challenge in this new task is related to the methods and techniques teams use and not the size of data set, We consider this data set to be large enough for the Grand Challenge and workshop in the first year of this task. The authors have also been in deliberations with the BBC regarding the licensing of the TV show \textit{Land Girls} for use in this data set, and the licensing of this series has been approved and will be available for subsequent years of this task. This is a 3-season / 15-episode series set in World War 2 about the lives of a group of women doing their part for Britain in the \textit{Women's Land Army} during the war.

\begin{center}
\begin{table}
\centering{
 \begin{tabular}{|c c c|} 
 \hline
 Movie & Genre & Duration\\ [0.5ex] 
 \hline\hline
 Honey & Romance&86\thinspace min \\
 \hline
 Let's Bring &&\\Back Sophie & Drama& 50\thinspace min\\
 \hline
 Nuclear Family & Drama & 28\thinspace min \\
 \hline
 Shooters & Drama & 41\thinspace min\\
 \hline
 Spiritual Contact &&\\ The Movie & Fantasy& 66\thinspace min\\
 \hline
 Super Hero & Fantasy & 18\thinspace min \\
 \hline
 The Adventures &&\\of Huckleberry Finn & Adventure & 106\thinspace min \\  
 \hline
 The Big Something & Comedy & 101\thinspace min\\
 \hline
 Time Expired & Comedy / Drama & 92\thinspace min \\
 \hline
 Valkaama & Adventure & 93\thinspace min \\ 
 \hline
  
\end{tabular}
\caption{The HLVU Dataset of open source movies}
\label{dataset} 
}
\end{table}
\end{center}

Due to the limited number of movies available for this task, the above list will be split 50-50 between a development set and test set. Final ground truths for movies selected as the development set will be made available to participants. Participating researchers will also be encouraged to procure their own suitable development sets and to share among other participating researchers.

\subsection{Human Annotation Framework}
\begin{figure*}[h]
  \centering
  \includegraphics[width=\linewidth, height=4.7in]{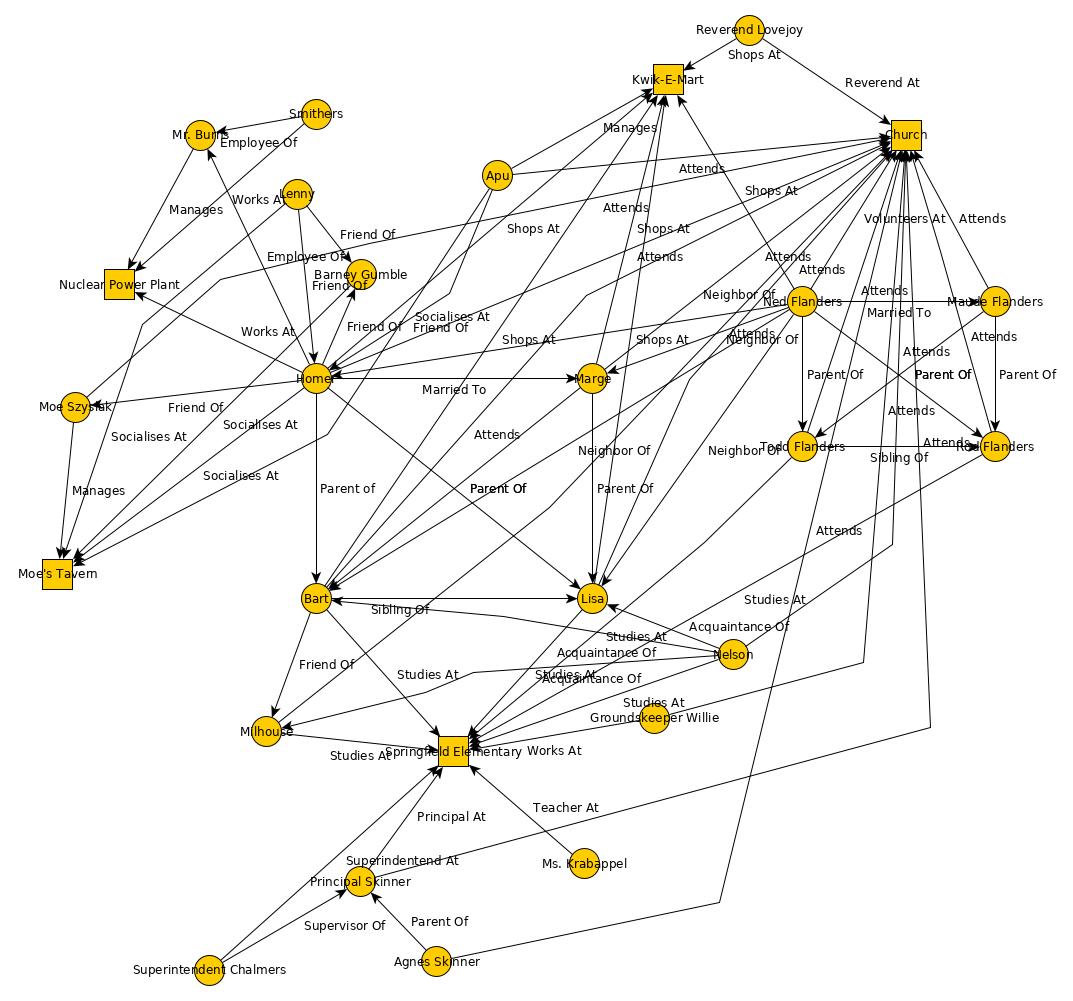}
  \caption{A sample knowledge graph depicting the world of \textit{The Simpsons}}
  \label{SimpsonsKG}
\end{figure*}

A group of human assessors will be recruited to provide manual annotations to the above HLVU dataset. These annotations will be the basis from which queries and ground-truth are generated. Each of the movies listed above will be watched at least twice by annotators. The first time is to familiarize themselves with the movie story line and the main actors. In the second time watching the movie they will be asked to use the yEd \cite{YED} graphing tool to develop a knowledge graph encompassing all of the actors, entities, relationships, and events. An example knowledge graph developed using this software, modelled on \textit{The Simpsons}, is shown in Figure \ref{SimpsonsKG}. In this example Knowledge Graph we map out the relations between the main characters of the show, in the way that generated Knowledge Graph's should map out the relations between main characters of the movie used in the dataset for this task.

Annotators will be provided with a primary list of possible relations between characters from which they must choose the most accurate relationship. Table \ref{relationships} provides some examples of the possible relations between characters and their inverse. If needed, they will be allowed to introduce new relationships as well.

\begin{figure*}[h]
  \centering
  \includegraphics[width=\linewidth, height=4in]{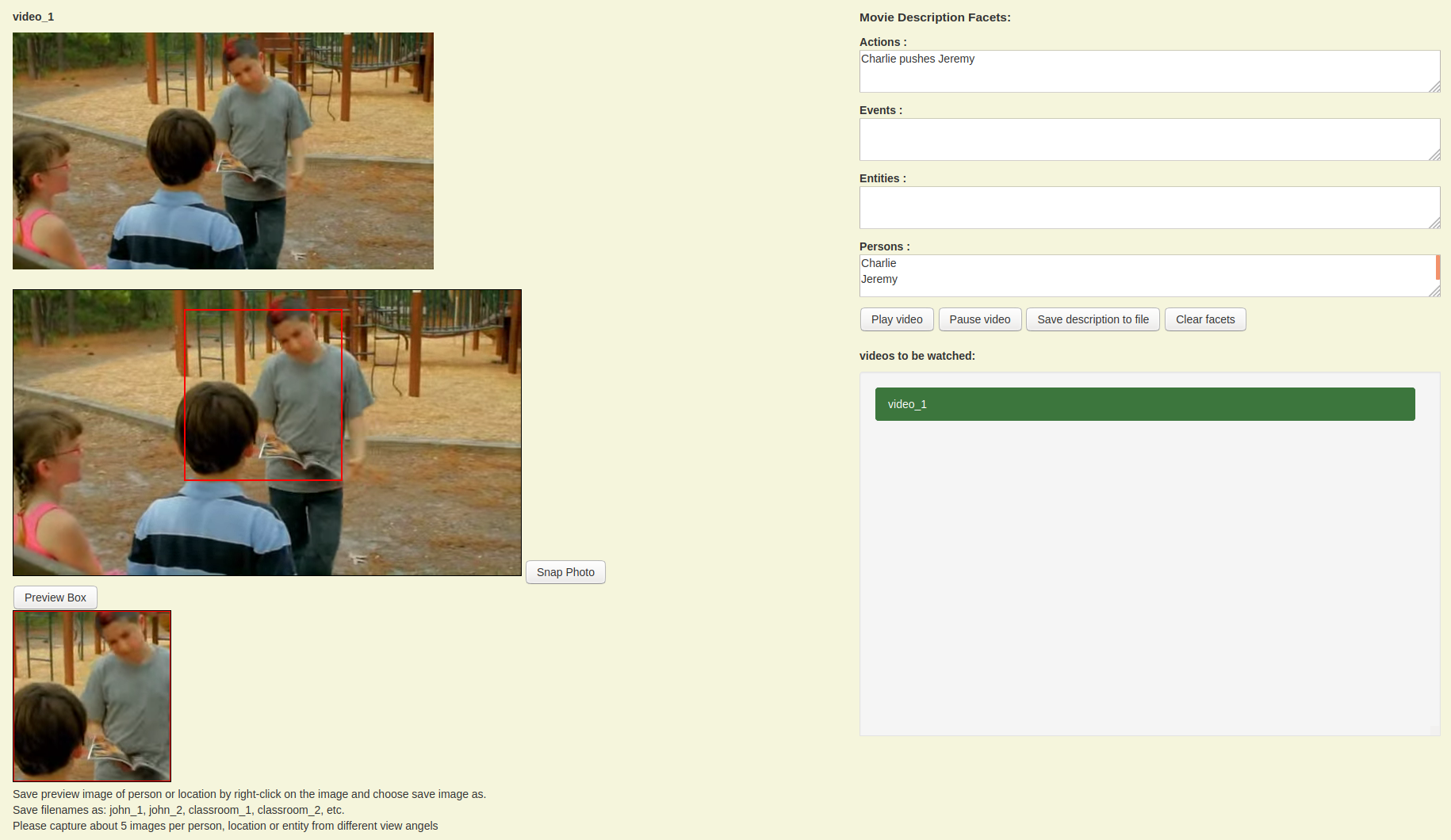}
  \caption{A screen shot of the annotation tool used for listing the defining actions and events for entities or relationships}
  \label{ant_tool}
\end{figure*}

In addition to the yEd graphing tool, a specially developed in-house annotation tool will also be used by annotators to save image captions of each of the actors and entities. This should also be used to list all the different actions which take place in the relationship between two people. Additionally this should be used to list key actions and events an individual has been involved in. These actions and events referred to here are not meant to be an exhaustive list and will be those actions and events that human annotators consider to be the key actions or events which may define the relationship between two individuals. In addition, the tool can be used to enable the annotators to document all the different events that happened during the movie and eventually enabling the development of a global ontology of events and actions relevant to the whole HLVU dataset. An example screen shot from using this annotation tool is shown in figure \ref{ant_tool}.

\begin{table}[h]
\begin{tabular}{|c |c |}
  \hline
  Relationship & Inverse Relationship\\ [0.5ex] 
  \hline\hline
  Child of & Parent of \\
  \hline
  Spouse of & Spouse of \\
  \hline
  Sibling of & Sibling of \\
  \hline
  Descendant of & Ancestor of \\
  \hline
  Friend of & Friend of \\
  \hline
  In Relationship With & In Relationship With \\
  \hline
  Ambivalent Of & Is Not Liked By \\
  \hline
  Employee of & Employer of \\
  \hline
  Attends & Attended By \\
  \hline
  Colleague of & Colleague of \\
  \hline
  Apprentice of & Mentor of \\
  \hline
  Student of & Teacher of \\
  \hline
  Supervisor of & Subordinate of \\
  \hline
  Superintendent at & Responsibility of \\
  \hline
\end{tabular}
\caption{Example Relationships Between Characters}
\label{relationships}
\end{table}

\subsection{Query Development}
The yEd graphing tool described in the previous section has the capability of exporting all entities and relationships to xgml or tgf format which can be used in the query development process to read all entities and relationships and develop different types of queries for evaluation purposes. The actions and events which may define the relationship between people may be derived from the in-house annotation tool described in the previous section as well. The following are a set of proposed query types which may be used to evaluate participating systems. Each query type tries to capture and test the systems for their comprehension of the tested movies from different points of view and levels of difficulty:

\subsubsection{A- Fill in the graph space:}
Fill in spaces in the Knowledge Graph (KG). Given the listed relationships, events or actions for certain nodes, where some nodes are replaced by variables X, Y, etc., solve for X, Y etc. Example: \textbf{X} Married To Marge. \textbf{X} Friend Of Lenny. \textbf{Y} Volunteers at Church. \textbf{Y} Neighbor Of \textbf{X}. Solution for \textbf{X} and \textbf{Y} in that case would be: \textbf{X} = Homer, \textbf{Y} = Ned Flanders. A more formal query example may look like the following xml block (asking who is the spouse of "Marge"):

\begin{lstlisting}
<Q.A>
  <Q.Id.1>
    <Subject>Person:Unknown_1</Subject>
    <Pred>Relation:Spouse_of</Pred>
    <Object>Person:Marge</Object>
  </Q.Id.1>
</Q.A>
\end{lstlisting}

\subsubsection{B- Question Answering:} 
This query type represents questions on the resulting KG, including actions and events, of the movies in the described dataset. For example, based on the Simpsons KG below, how many children does Marge have? By counting the 'Parent Of' edges from Marge to other nodes, we can see that Marge has two children, Bart and Lisa. Other possible questions may be for example: What does Ms. Krabappel do for a living? and the answer can be a multiple choice from the set of provided entity relationships. 
A query of this type may look like the following xml block:
\begin{lstlisting}
<Q.B>
  <Q.Id.1>
    <Subject>Person:Ms. Krabappel</Subject>
    <Pred>Relation:Unknown_1</Pred>
    <Object>Location:Springfield Elementary</Object>
    <Ans_1>Relation: X</Ans_1>
    <Ans_2>Relation: Y</Ans_2>
     .
     .
    <Ans_n>Relation: Z</Ans_n>
  </Q.Id.1>
</Q.B>
\end{lstlisting}

\begin{figure}[h]
  \centering
  \includegraphics[width=\linewidth]{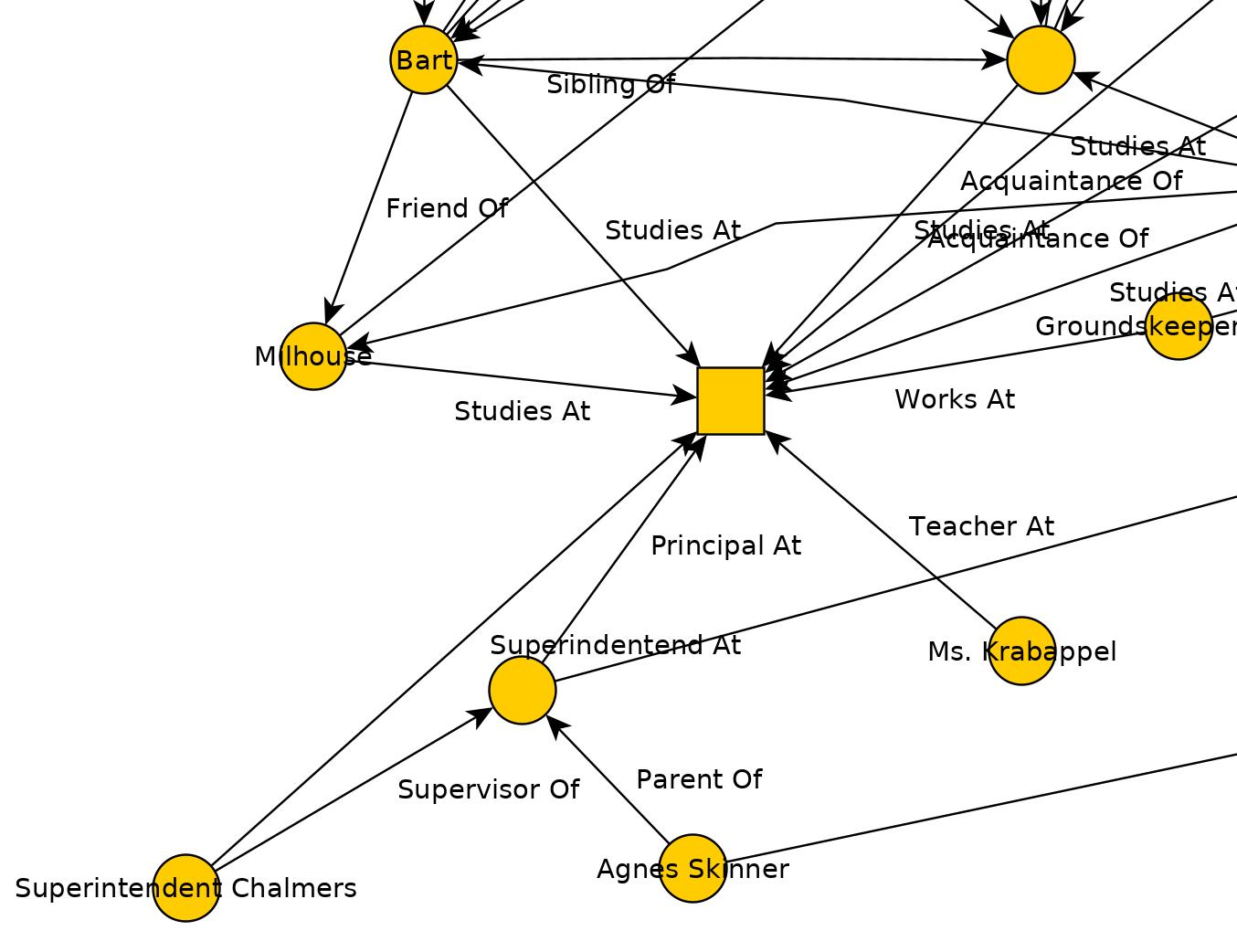}
  \caption{Visible region of graph with three blank nodes to be filled in.}
  \label{vis_region}
\end{figure}

\subsubsection{C- Relations between characters:}
How is character X related to character Y ? This query type question asks participants about all routes through the KG from one person to another. The main objective of this query type is to test the quality of the established KG. If the system managed to build a representative KG reflecting the real story line of the movie, then it should be able to return back all valid paths, including the shortest, between characters (i.e. how they are related to each other).
For an example, looking at the Simpsons KG in figure \ref{SimpsonsKG}, what would be the shortest route for Superintendent Chalmers, the left-bottom most node, to deliver a message to Lenny, in the top left hand corner of the knowledge graph?
To answer this question we would follow the edges connecting each node and trace all possible paths from Superintendent Chalmers to Lenny, settling on the three shortest routes: (1) Superintendent Chalmers is supervisor to Principal Skinner, Principal Skinner attends Church, Church is also attended by Homer, Homer is in turn Lenny's friend. (2) Superintendent Chalmers is Superintendent at Springfield Elementary. Springfield Elementary is studied at by Bart. Bart is Child of Homer. Homer is friend of Lenny. (3) Superintendent Chalmers is Superintendent at Springfield Elementary. Springfield Elementary is studied at by Lisa. Lisa is Child of Homer. Homer is friend of Lenny. The below is the formal query representation for the above example:
\begin{lstlisting}
<Q.C>
  <Q.Id.1>
    <Source>Person:Superintendent Chalmers</Source>
    <Target>Person:Lenny</Target>
  </Q.Id.1>
</Q.C>
\end{lstlisting}

\subsection{Evaluation Framework}
Using the annotation and graphing tools described above, we can develop evaluation software to read as inputs the ground truth provided by these annotation and graphing tools, in addition to reading as input the submissions of participating teams, automatically evaluating each team's submissions, scoring and ranking results based on the selected evaluation metric. We should note here that systems will be given a set of image and/or video examples for the different actors and entities including important locations, each with a name Id. In addition, the final ontology of relationships, events, and actions will also be given so that systems can align the different query mentions of people, relations, entities, etc and return results from the given ontology classes as well. For illustration purposes, the following is a system response for each of the three query types and a proposed metric to score submissions:

\subsubsection{A- Fill in the graph space:}
Given the partial graph query illustrated in figure \ref{vis_region}, systems are asked to fill in the three blank spaces in the KG. The query itself will be given in xml format as illustrated in the query development section above. The ground truth for this example is: Lisa is sibling of Bart. Springfield Elementary is where Milhouse and Bart study at, and Principal Skinner is the principal. An example response by a system is shown below giving a set of answers for each unknown graph node being asked about in addition to a confidence score. Results will be treated as ranked list of result items per each unknown variable and the Reciprocal Rank score will be calculated per unknown variable and Mean Reciprocal Rank (MRR) per query \cite{voorhees1999trec}.

\begin{lstlisting}
<Q.A>
  <Q.Id.1>
    <Person:Unknown_1><Ans_1><Confidence>
    <Person:Unknown_1><Ans_2><Confidence>
     ...
     ...
    <Person:Unknown_2><Ans_n><Confidence>
  </Q.Id.1>
</Q.A>
\end{lstlisting}

The MRR measurement (Eq. \ref{q1_eqn}) evaluates systems which return a ranked list of answers to questions. For this reason we consider it the most appropriate evaluation measurement for use in this query type. It is the average of the reciprocal ranks of results for a group of queries. For example, in three ranked lists submitted to answer three unknown variables in a query, if the answer to the first query is ranked second in the list, the answer to the second query is ranked first in the list and the answer to the third is ranked fourth, this gives \( \frac{1}{2} \), 1, and \( \frac{1}{4} \). This averages to give an MRR score of \( \frac{7}{12} \).

\begin{equation} \label{q1_eqn}
MRR = \frac{1}{Q}\sum_{i=1}^{Q}\frac{1}{RANK {i}}
\end{equation}



\subsubsection{B- Question Answering:} 
Multiple choice questions will be provided for participants to answer which will be based on the KG. For an example on the format and types of questions asked to participants please see the Query Development section above. The below is an example of a system response and the evaluation metric (Eq. \ref{q2_eqn}) proposed for this query type.

\begin{lstlisting}
<Q.B>
  <Q.Id.1>
    <Answer>Relation: Z</Answer>
  </Q.Id.1>
</Q.B>
\end{lstlisting}

\begin{equation} \label{q2_eqn}
Score = \frac{Correct Answers}{Total Questions}
\end{equation}

\subsubsection{C- Relations between characters:}

In this query type, systems are asked to submit all valid paths form a source node to another target node with the goal of maximizing recall and precision. NIST will first evaluate whether each path is a valid path (i.e the submitted order of nodes and edges leads to a path from the source person to the target person of the query) and report the recall, precision and F1 measures \cite{van1979information}:

\begin{equation} \label{q3_eqn}
Recall = \frac{Number of Submitted Valid Paths}{Number of Valid GT Paths}
\end{equation}

\begin{equation} \label{q3_eqn}
Precision = \frac{Number of Submitted Valid Paths}{Number of Submitted Paths}
\end{equation}

\begin{equation} \label{q3_eqn}
F1\_Score = 2*\frac{Precision * Recall}{Precision + Recall}
\end{equation}

Below is an example of a system response to a query asking for all paths from the Simpsons characters Superintendent Chalmers to Lenny:

\begin{lstlisting}
<Q.C>
  <Q.Id.1>
    <path=1>
    <Source>Person:Superintendent Chalmers</Source>
    <edge>Relation:Superintendent_At</edge>
    <node>Entity:Springfield Elementary</node>
    <edge>Relation:Studied_At_By</edge>
    <node>Person:Bart</node>
    .
    .
    <Target>Person:Lenny</Target>
    </path>
    <path=2>
    <Source>Person:Superintendent Chalmers</Source>
    .
    .
    </path>
    <path=3>
    .
    .
    </path>
  </Q.Id.1>
</Q.C>
\end{lstlisting}

The evaluation framework described above will provide a way to accurately and automatically evaluate the Knowledge Graphs produced by participating researchers. This allows for consistent scoring methods to be applied to this task which increases confidence in results and allows for easy expansion of the task and number of participants.

\section{Conclusions and Future Directions}
In this paper we discussed a proposed new evaluation benchmark and associated pilot dataset (HLVU) to promote research in holistic and deep video understanding. The evaluation paradigm aims for long duration self contained story lines in movies and targets automatic systems to eventually build complete knowledge graphs representing the movie characters, relationships, and key actions and events associated with them. The different query types presented aim to test the current automatic computer vision systems' capabilities and measure if they can "understand" video story lines the way humans do. The future plans for this evaluation benchmark is to organize an ACM Multimedia Grand Challenge and a workshop at ACM ICMI to host the evaluation campaign encouraging researchers to explore the dataset, discuss challenges and lessons learned trying to solve the associated problems with deep video understanding of the small world of movies. Also, securing more long term stable and large-scale dataset of licensed or open source movies including soap opera series of closed world story lines is a major future plan to continue measuring progress and the state of the art in this new domain.


\begin{acks}
The authors would like to thank the Information Technology Laboratory at NIST for sponsoring the work presented in this paper.
\end{acks}


\emph{Disclaimer: Certain commercial entities, equipment, or
materials may be identified in this document in order to describe an
experimental procedure or concept adequately. Such identification is
not intended to imply recommendation or endorsement by the National
Institute of Standards and Technology, nor is it intended to imply 
that the entities, materials, or equipment are necessarily the best 
available for the purpose. The views and conclusions contained herein 
are those of the authors and~should not be interpreted as necessarily
representing the official policies or endorsements, either expressed or 
implied, of NIST, or the U.S. Government.}

\bibliographystyle{ACM-Reference-Format}
\bibliography{acmart}



\end{document}